\documentclass[conference,a4paper]{APSIPA2025}
\usepackage{amsmath}
\usepackage{graphicx}
\usepackage{multirow}
\usepackage{threeparttable}
\usepackage{amsmath,amssymb}
\usepackage{here}
\usepackage{url}
\usepackage{comment}
\usepackage{multirow}
\usepackage{tabularx}
\usepackage{color}
\usepackage{xspace}
\usepackage{dsfont}
\usepackage{xcolor}
\usepackage{amssymb}
\usepackage{bm}
\usepackage{pdfpages}
\usepackage{cite}
\usepackage{booktabs}

\usepackage{geometry}
\begin{document}

\title{Joint Modeling of Big Five and HEXACO for Multimodal Apparent Personality-trait Recognition}

\author{
\authorblockN{
Ryo Masumura, 
Shota Orihashi, 
Mana Ihori, 
Tomohiro Tanaka, 
Naoki Makishima, 
Taiga Yamane, \\
Naotaka Kawata, 
Satoshi Suzuki, 
Taichi Katayama
}

\authorblockA{
NTT, Inc., Japan \\
E-mail: ryo.masumura@ntt.com}
}


\maketitle

\begin{abstract}
This paper proposes a joint modeling method of the Big Five, which has long been studied, and HEXACO, which has recently attracted attention in psychology, for automatically recognizing apparent personality traits from multimodal human behavior. Most previous studies have used the Big Five for multimodal apparent personality-trait recognition. However, no study has focused on apparent HEXACO which can evaluate an {\it Honesty-Humility} trait related to displaced aggression and vengefulness, social-dominance orientation, etc. In addition, the relationships between the Big Five and HEXACO when modeled by machine learning have not been clarified. We expect awareness of multimodal human behavior to improve by considering these relationships. The key advance of our proposed method is to optimize jointly recognizing the Big Five and HEXACO. Experiments using a self-introduction video dataset demonstrate that the proposed method can effectively recognize the Big Five and HEXACO. 
\end{abstract}

\section{Introduction}
Recognizing people's personality traits has been a central topic in the psychological and engineering fields. Two types of personality traits have been considered; self-assessed and apparent perceived by observers. In psychology, personality traits are measured through questionnaire-based personality tests for both the self-assessed and apparent personality traits. While the personality test results for self-assessed personality traits can be attained from one self-trial, those for apparent personality traits need to be judged by many other people. To recognize the apparent personality traits without the help of people other than oneself, researchers in the engineering field have studied multimodal apparent personality-trait recognition in which apparent personality traits are automatically recognized from multimodal human behavior using machine learning \cite{mehta_2020,zhao_2022,esalante_ieee2022,ilmini_2024}.

\begin{table*}[t!]
  \centering
  \caption{A 60-item HEXACO questionnaire.}
  \begin{tabular}{l|c|l} \toprule
id & key & question  \\ \midrule 
1. & O- &  He/she would be quite bored by a visit to an art gallery. \\
2. & C+ &  He/she plans ahead and organizes things, to avoid scrambling at the last minute. \\
3. & A+ &  He/she rarely holds a grudge, even against people who have badly wronged him/her. \\
4. & X+ &  He/she feels reasonably satisfied with himself/herself overall. \\
5. & E+ &  He/shewould feel afraid if he/she had to travel in bad weather conditions. \\
6. & H+ &  He/she wouldn't use flattery to get a raise or promotion at work, even if he/she thought it would succeed.\\
... & ... & ... \\ 
55. & O- &  He/she finds it boring to discuss philosophy. \\
56. & C- & He/she prefers to do whatever comes to mind, rather than stick to a plan. \\
57. & A- & When people tell him/her that he/she is wrong, his/her first reaction is to argue with them. \\
58. & X+ & When he/she is in a group of people, he/she is often the one who speaks on behalf of the group.\\
59. & E- &He/she remains unemotional even in situations where most people get very sentimental.\\
60. & H- &  He/she’d be tempted to use counterfeit money, if he/she were sure he/she could get away with it. \\ \bottomrule
    \end{tabular}
\end{table*} 
Many modeling methods for multimodal personality-trait recognition have been studied. Deep-learning-based methods for learning effective representations from multimodal human behavior without introducing hand-crafted features are now widely used \cite{gucluturk_eccv2016,gobova_ieee2018,kampman_2018,principi_ieee2021,aslan_2021}. With these methods, personality traits are estimated by integrating speech, visual, and text information exploited from human behavior. When modeling apparent personality traits, most studies modeled to recognize the Big Five personality traits of {\it Openness}, {\it Conscientiousness}, {\it Extraversion}, {\it Agreeableness}, and {\it Neuroticism} \cite{goldberg_1990,mccrae_1992}. However, no study has focused on recognizing apparent personality traits other than the Big Five. This is because most datasets were developed for measuring the Big Five \cite{zhao_2022}. 

In this study, we focus on the HEXACO traits \cite{lee_hexaco_2004,ashton_hexaco_2007} supported by recent theoretical and empirical studies on alternatives to the Big Five. HEXACO is a six-factor framework that includes {\it Honesty-Humility} and variants of the Big Five traits, i.e., {\it Emotionality}, {\it Extraversion}, {\it Agreeableness}, {\it Conscientiousness}, and {\it Openness}. It has been investigated that {\it Honesty-Humility} is strongly negatively correlated with a variety of factors (e.g., displaced aggression and vengefulness \cite{lee_personality_2012}, social-dominance orientation \cite{leone_personality_2012}, and workplace misconduct \cite{pletzer_personality_2019}) and has little correlation with the Big Five traits, so it would be worthwhile to automatically recognize the apparent HEXACO personality traits from multimodal human behavior. Note that there was one trial that examined self-reported HEXACO traits from social-media text posts \cite{sinha_wassa_2015}, but inferring apparent observer-perceiving HEXACO traits from multimodal human behavior has not been investigated. In addition, the relationships between the Big Five and HEXACO when modeled by machine learning have not been clarified, although their relationships have been analyzed from many psychological aspects. For example, characteristics other than {\it Honesty-Humility} in HEXACO are closely related to the corresponding characteristics in the Big Five \cite{ashton_personality_2005,vries_personality_2013}. It has also been shown that {\it Honesty-Humility} is partially related to {\it Agreeableness} of the Big Five \cite{howard_personality_2020}. By modeling multimodal personality-trait recognition that can take into account these relationships, we expect to promote robustness to being aware of various multimodal human behaviors.

To explicitly consider the relationships between the Big Five and HEXACO, we propose a joint-modeling method of the Big Five and HEXACO for multimodal apparent personality-trait recognition. Our proposed method simultaneously optimizes recognizing the Big Five and HEXACO from multimodal audio-video information. We model them using a multimodal-transformer architecture \cite{liao_ieee2023} to increase the awareness of multimodal human behavior in the Big Five and HEXACO. For this modeling, we extend a existing self-introduction video dataset \cite{masumura_aaai2025} by assigning not only the Big Five and HEXACO. Our dataset consists of 50 Big Five questionnaire items \cite{goldberg_1993,apple_2012} and 60 HEXACO questionnaire items \cite{ashton_hexaco60_2007} collected from five observers of over 10,000 self-introduction videos. In experiments using the dataset, we show that joint modeling can improve the recognition performance of Big Five and HEXACO compared with individual modeling.

Our contributions are summarizes as follows.
\begin{itemize}
\item This paper is the first to examine multimodal apparent personality-trait recognition involving HEXACO.
\item This paper provides a joint modeling method of the Big Five and HEXACO, which yields the improved recognition performance of both traits.
\item This paper is the first to investigate the relationships between the Big Five and other personality traits, i.e., HEXACO, in multimodal apparent personality-trait recognition.
\item This paper presents a self-introduction video dataset to which the Big Five and HEXACO traits are jointly assigned by others. 
\end{itemize}

\section{Dataset}
This section details our self-introduction video dataset.

\subsection{Self-introduction Videos}
We extended a existing self-introduction video dataset \cite{masumura_aaai2025} by assigning not only the Big Five and HEXACO. The dataset includes 10,100 self-introduction videos collected from 1,010 participants. The following interview items are on the theme of self-introduction. 
``Please tell us about your hobbies.'' 
``Please tell us about your favorite food.'' 
``Please tell us about your favorite celebrity.'' 
``Please tell us about the tourist spots that you are glad you visited.'' 
``Please tell us about your most impressive childhood memories.'' 
``Please tell us about some interesting people you have met.'' 
``Please tell us about your favorite season.'' 
``Please tell us about the place you would like to visit.'' 
``Please tell us about something you would like to try.'' 
``Please tell us about something you are not good at''. 
Ten videos were recorded from each participant, who were all Japanese. The recorded videos are composed of about 12,395 min of recordings, and the average duration of each video is 73.6 s. The maximum and minimum duration of all videos are 102.1 and 59.1 s, respectively. All videos were recorded using Zoom on laptop PCs. We recorded the videos at 25 fps in 1280 × 720 resolution. Camera views were frontal, and we recorded the upper part of the body. The audio was recorded at 16 kHz. We split the dataset into a training dataset containing 9,030 videos recorded from 903 participants, validation dataset containing 500 videos recorded from 50 participants, and test dataset containing 570 videos recorded from the remaining 57 participants.

\subsection{Annotations of Big Five and HEXACO}
All recorded videos were annotated with apparent personality traits. We used the Big Five \cite{goldberg_1990,mccrae_1992} (\textit{Openness}, \textit{Conscientiousness}, \textit{Extraversion}, \textit{Agreeableness}, and \textit{Neuroticism}) and HEXACO \cite{lee_hexaco_2004,ashton_hexaco_2007} (\textit{Honesty-Humility}, \textit{Emotionality}, \textit{Extraversion}, \textit{Agreeableness}, \textit{Conscientiousness}, and \textit{Openness}) for the apparent personality traits. To annotate people’s apparent personality traits, we recruited 200 observers who did not know the 1,010 participants. We used a 50-item Big Five questionnaire \cite{goldberg_1993,apple_2012} and 60-item HEXACO questionnaire \cite{ashton_hexaco60_2007}. The videos in the training and validation datasets were scored by five randomly selected observers and those in the test dataset were scored by ten randomly selected observers . In the test dataset, five annotations assigned ground-truth information, and the other five conducted human evaluation. Each observer watched each recorded video two or three times and answered the questionnaire. We used a five-point scale for scoring. Table 1 shows the 12 items in the 60-item HEXACO questionnaire. Each key in Table 1 represents which personality traits it pertains to. ``H'', ``E'', ``'X'', ``A'', ``C'' and ``O'' represent \textit{Honesty-Humility}, \textit{Emotionality}, \textit{Extraversion}, \textit{Agreeableness}, \textit{Conscientiousness}, and \textit{Openness}, respectively. For ``+'' keyed items, the response ``Very Inaccurate'' is assigned a value of 1, ``Moderately Inaccurate'' a value of 2, ``Neither Inaccurate nor Accurate'' a 3, ``Moderately Accurate'' a 4, and ``Very Accurate'' a value of 5. For ``-'' keyed items, the response ``Very Inaccurate'' is assigned a value of 5, ``Moderately Inaccurate'' a value of 4, ``Neither Inaccurate nor Accurate'' a 3, ``Moderately Accurate'' a 2, and ``Very Accurate'' a value of 1. Note that the annotators were instructed to avoid assigning ``Neither Inaccurate nor Accurate'' as much as possible.
Once scores are assigned for all of the items in the scale, all the values are averaged to obtain a total scale score. Figures 1 and 2 respectively show the histograms of the annotated Big Five and HEXACO personality traits of our recorded videos. The scores of individual personality traits are in the range of [1, 5]. Note that these scores are normalized in the range of [0, 1] when using deep-learning-based modeling methods.

\begin{figure}[t!]
  \begin{center}
    \includegraphics[width=85mm]{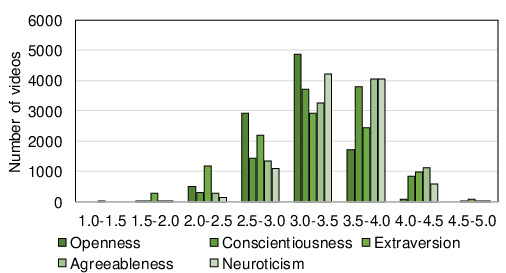}
  \end{center}
  \caption{The histograms of the annotated Big Five}
\end{figure}

\begin{figure}[t!]
  \begin{center}
    \includegraphics[width=85mm]{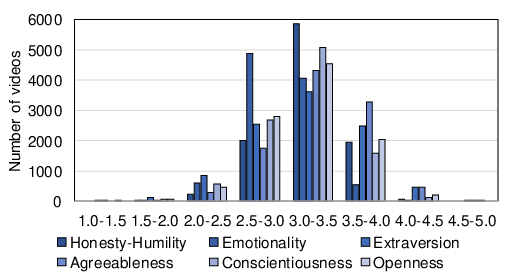}
  \end{center}
  \caption{The histograms of the annotated HEXACO}
\end{figure}

\section{Joint Modeling of Big Five and HEXACO with Multimodal Transformer}
This section details a joint modeling method of the Big Five and HEXACO.

\subsection{Definitions}
In this task, the Big Five scores $\hat{\bm{y}} = [\hat{y}_1,\cdots,\hat{y}_5]^{\top}$ and HEXACO scores $\hat{\bm{z}} = [\hat{z}_1,\cdots,\hat{z}_6]^{\top}$ are jointly estimated from an audio-visual video input, which is represented as audio features $\bm{S}$ and their corresponding visual features $\bm{U}$. Audio features are generally extracted from speech information, and visual features are extracted from human RGB images. When modeling multimodal fine-grained apparent-personality-trait recognition, $\hat{\bm{y}}$ and $\hat{\bm{z}}$ are estimated using 
\begin{equation}
 \{\hat{\bm{y}}, \hat{\bm{z}}\} = {\mathcal F}(\bm{S},\bm{U}; \bm{\Theta}) ,
\end{equation}
where ${\mathcal F}(\cdot)$ is the model function and $\bm{\Theta}$ represents the trainable model-parameter set. In addition, an automatic speech recognition (ASR) system can be used to convert the $\bm{S}$ into text $\bm{W}$.

\subsection{Joint Modeling}
Our proposed method uses a multimodal transformer architecture to effectively capture multimodal information. The advantage of this is that different types of features can be handled with the same input method. The architecture consists of four encoders: audio, text, visual, and multimodal. Figure 3 shows the architecture. The audio encoder converts audio features $\bm{S}$ into audio representations $\bm{A}$, the text encoder converts text $\bm{W}$ into text representations $\bm{T}$, and the visual encoder converts visual features $\bm{U}$ into visual representations $\bm{V}$. 

\begin{figure}[t!]
  \begin{center}
    \includegraphics[width=85mm]{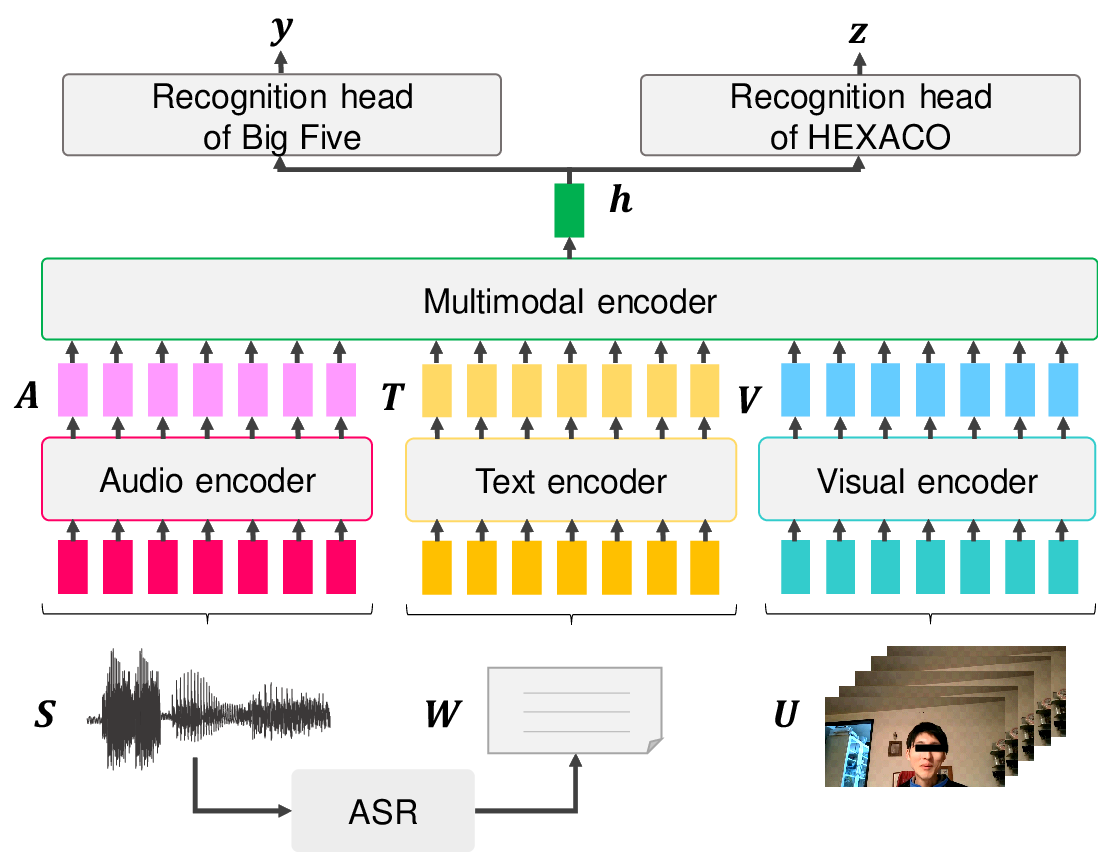}
  \end{center}
  \caption{Joint modeling of Big Five and HEXACO with multimodal transformer}
\end{figure} 
The multimodal encoder handles cross-modal interactions of outputs from the audio, text, and visual encoders. The inputs for the multimodal encoder are 
\begin{equation}
\bm{H}_{0} = 
\begin{cases}
{\rm TemporalConcat}(\bm{A},\bm{T},\bm{V})  & \text{if ASR is performed}, \\
{\rm TemporalConcat}(\bm{A},\bm{V}) & \text{else},
\end{cases}
\end{equation}
\begin{equation}
 \bm{H}^\prime_{0} = {\rm AddSegment}(\bm{H}_0; \bm{\theta}_{\rm segment}) ,
 \end{equation}
where ${\rm TemporalConcat}()$ is a function that concatenates inputs on the temporal axis, ${\rm AddSegment}()$ is a function that adds a continuous vector in which modal-specific segment information is embedded to distinguish the concatenated vectors, and $\bm{\theta}_{\rm segment} \in \bm{\Theta}$ are the trainable parameters. We obtain hidden vectors $\bm{H}$ by
\begin{equation}
 \bm{H} = {\rm TransformerEnc}(\bm{H}^\prime_0; \bm{\theta}_{\rm multi}) ,
 \end{equation}
where ${\rm TransformerEnc}()$ is a function of the transformer encoder blocks \cite{vaswani_nips2017} and $\bm{\theta}_{\rm multi} \in \bm{\Theta}$ are the trainable parameters of the multimodal encoder. Note that the length of $\bm{H}$ changes depending on the inputs. 

Attentive pooling converts variable length $\bm{H}$ into a fixed size vector. The fixed vector is obtained by 
\begin{equation}
 \bm{h} = {\rm AttentivePooling}(\bm{H}; \bm{\theta}_{\rm pool}) ,     
\end{equation}
where $\bm{\theta}_{\rm pool} \in \bm{\Theta}$ are the trainable parameters of attentive pooling, and ${\rm AttentivePooling}()$ is the attentive-pooling function.

This model jointly estimates the Big Five and HEXACO scores by providing two prediction heads calculated as
\begin{eqnarray}
\hat{\bm{z}} & = & {\rm Sigmoid}(\bm{h}; \bm{\theta}_{\rm head}^{\rm z}) , \\
\hat{\bm{y}} & = & {\rm Sigmoid}(\bm{h}; \bm{\theta}_{\rm head}^{\rm y}) , 
\end{eqnarray}
where $\{\bm{\theta}_{\rm head}^{\rm z}, \bm{\theta}_{\rm head}^{\rm y}\} \in \bm{\Theta}$ are the trainable parameters.

\subsection{Training}
To train $\bm{\Theta}$, we use a dataset of audio-visual video input, which is expressed as
\begin{equation}
{\mathcal D} = \{(\bm{S}^1, \bm{U}^1, \bm{y}^1, \bm{z}^1),\cdots, (\bm{S}^{|\mathcal D|}, \bm{U}^{|\mathcal D|}, \bm{y}^{|\mathcal D|},\bm{z}^{|\mathcal D|})\}.
\end{equation}

Our joint model is trained with the mean absolute error loss between the ground-truth Big Five and estimated Big Five, and  the mean absolute error loss between the ground-truth HEXACO and estimated HEXACO as
\begin{equation}
{\mathcal L} = \frac{1}{|\mathcal D|} \sum_{d = 1}^{|\mathcal D|} |\hat{\bm{y}}^d - \bm{y}^d| +  \frac{1}{|\mathcal D|} \sum_{d = 1}^{|\mathcal D|} |\hat{\bm{z}}^d - \bm{z}^d| .
\end{equation}
By taking into account the relationships between the Big Five and HEXACO, we expect to promote robustness to being aware of various multimodal human behaviors.

\section{Experiments}
We used our dataset in the following experiments. We verified the effectiveness of our proposed joint-modeling method. We also investigated the relationships between the Big Five and HEXACO in multimodal apparent personality-trait recognition.

\begin{table*}[ht!]
  \centering
  \caption{Experimental results of recognizing Big Five in terms of Pearson’s correlation coefficient (Corr.) and accuracy (Acc.)}
   \vspace{-1em}
  \begin{tabular}{l|c|cccccccccc} \toprule
       Modeling & Input & \multicolumn{2}{c}{\textit{Openness}} & \multicolumn{2}{c}{\textit{Conscientiousness}} & \multicolumn{2}{c}{\textit{Extraversion}} & \multicolumn{2}{c}{\textit{Agreeableness}} & \multicolumn{2}{c}{\textit{Neuroticism}} \\
    method & modals & Corr. & Acc. & Corr. & Acc. & Corr. & Acc. & Corr. & Acc. & Corr. & Acc. \\ \midrule
    Big Five model & Audio & 0.493 & 93.9 & 0.604 &  93.2 & 0.647 & 91.2 &  0.572 &  92.3 & 0.473 & 93.5 \\  
    Joint model & Audio & {\bf 0.542} & {\bf 94.4} & {\bf 0.614} & {\bf 93.3} & {\bf 0.707} & {\bf 91.6} & {\bf 0.576} & {\bf 93.4} & {\bf 0.530} & {\bf 93.8}  \\ \midrule
    Big Five model & Visual & {\bf 0.233} & {\bf 93.1} & 0.310 & 90.8 &  0.264 & 86.4 & 0.433 & 92.4 & 0.233 &  93.1  \\   
    Joint model & Visual & 0.228 & 92.9 & {\bf 0.332} & {\bf 91.2} & {\bf 0.315} & {\bf 87.2} & {\bf 0.452} & {\bf 92.6} & {\bf 0.286} & {\bf 93.3} \\ \midrule    
    Big Five model & Audio, Visual & 0.544 & 94.4 &  0.604 & {\bf 93.5} & 0.735 & 91.0 & 0.615  & 92.6 & 0.532 & 94.0  \\
    Joint model & Audio, Visual & {\bf 0.557} & {\bf 94.5} & {\bf 0.617} & 93.3 & {\bf 0.743} & {\bf 92.0} & {\bf 0.628} & {\bf 93.8} & {\bf 0.538} & {\bf 94.2}  \\ \midrule    
    Big Five model & Audio, Visual, Text & 0.585 & 94.6 & 0.675 & 93.8 & 0.752 &  92.4 & 0.617 & 92.7 & {\bf 0.586} & 94.1  \\
    Joint model & Audio, Visual, Text & {\bf 0.595} & {\bf 94.8} & {\bf 0.686} & {\bf 93.9} & {\bf 0.757} &  {\bf 92.6} & {\bf 0.657} & {\bf 94.0} & {\bf 0.586} & {\bf 94.2}  \\ \midrule
    \multicolumn{2}{c|}{\textit{Human evaluation}} & 0.544 & 92.9 & 0.668 & 92.7 & 0.770 & 91.7 & 0.645& 92.4& 0.532 & 92.1 \\ \bottomrule
    \end{tabular}
\end{table*}

\begin{table*}[ht!]
  \centering
  \caption{Experimental results of recognizing HEXACO in terms of  Pearson’s correlation coefficient (Corr.) and accuracy (Acc.)}
  \begin{tabular}{l|c|cccccccccccc} \toprule
       Modeling & Input & \multicolumn{2}{c}{\textit{Honesty-Humility}} & \multicolumn{2}{c}{\textit{Emotionality}} & \multicolumn{2}{c}{\textit{Extraversion}} & \multicolumn{2}{c}{\textit{Agreeableness}} & \multicolumn{2}{c}{\textit{Conscientiousness}} & \multicolumn{2}{c}{\textit{Openness}} \\
    method & modals & Corr. & Acc. & Corr. & Acc. & Corr. & Acc. & Corr. & Acc. & Corr. & Acc. & Corr. & Acc. \\ \midrule
    HEXACO model & Audio & 0.468 & 95.1 & 0.626 & 95.3 & 0.616 & 92.7 & 0.468 & 94.0 & 0.546 & 93.8 & {\bf 0.456} & 93.7 \\  
    Joint model & Audio & {\bf 0.482} & {\bf 95.2} & {\bf 0.639} & {\bf 95.6} & {\bf 0.660} & {\bf 92.9} & {\bf 0.469} & {\bf 94.0} & {\bf 0.549} & {\bf 94.1} & 0.454 & {\bf 93.7} \\  \midrule
    HEXACO model & Visual & {\bf 0.220} & 94.5 & 0.495 & 94.7 & 0.305 & 89.9 & 0.443 & 93.6 & 0.204 & 92.5 & 0.278 & 93.0 \\  
    Joint model & Visual & 0.214 & {\bf 94.5} & {\bf 0.502} & {\bf 94.8} & {\bf 0.320} & {\bf 90.4} & {\bf 0.454} & {\bf 93.7} & 0.198 & {\bf 92.8} & {\bf 0.290} & {\bf 93.3} \\  \midrule
    HEXACO model & Audio, Visual & 0.477 & 95.1 & 0.627 & 95.2 &  0.681 & 93.0 & 0.551 & 94.3 & 0.541 & {\bf 94.0} & 0.491 & 93.0 \\  
    Joint model & Audio, Visual & {\bf 0.480} & {\bf 95.2} & {\bf 0.635} & {\bf 95.4} & {\bf 0.691} & 92.9 & {\bf 0.568} & 94.2 & {\bf 0.547} & {\bf 94.0} & {\bf 0.504} & {\bf 93.8} \\  \midrule
    HEXACO model & Audio, Visual, Text & 0.492 & 94.6 & 0.651 & 95.3 & 0.693 & 93.1 & 0.570 & 93.6 & 0.559 & 94.0 & 0.594 & {\bf 94.4} \\  
    Joint model & Audio, Visual, Text & {\bf 0.504} & {\bf 95.2} & {\bf 0.645} & {\bf 95.6} & {\bf 0.707}  & {\bf 93.2} & {\bf 0.576} & {\bf 94.3} & {\bf 0.579} & {\bf 94.2} & {\bf 0.608} & {\bf 94.4} \\  \midrule
    \multicolumn{2}{c|}{\textit{Human evaluation}} & 0.401 & 92.6 & 0.497 & 93.3 & 0.744 & 93.1 & 0.555& 92.5& 0.592 & 92.8 & 0.536 & 92.3 \\ \bottomrule
    \end{tabular}
\end{table*}

\subsection{Setups}
In our evaluation, we constructed two task-specific models, i.e., Big Five model and HEXACO model, and a joint model using a multimodal transformer architecture.

We carried out pre-processing to extract audio and visual features from video input. We extracted 80 log Mel-scale filterbank coefficients for the acoustic features, and the frame shift was 10 ms. Face regions in each input frame were detected with CenterNet \cite{zhou_2019} trained on the Wider Face dataset \cite{yang_cvpr2016} for the visual features. The face images were cropped and resized to 128\,$\times$\,128, and down-sampled to 3 fps. We converted the audio features into text using a transformer-based end-to-end automatic-speech-recognition (ASR) system trained with 20K hrs of Japanese speech. The configuration was as follows. For the audio encoder, audio features passed two convolution and max-pooling layers with a stride of 2, so we down-sampled them to $1/4$ along with the time axis. We stacked four transformer-encoder blocks. For the visual encoder, the convolutional-neural-network function was composed of the MobileNetV3 architecture \cite{howard_iccv2019}, and two transformer encoder blocks were additionally stacked. We stacked six transformer-encoder blocks for the text encoder and two transformer-encoder blocks for the multimodal encoder. For each encoder, the dimensions of the output continuous representations were set to 256, dimensions of the inner outputs were set to 1024, and number of heads in the multi-head attentions was set to 4. Swish activation was used for these encoders. For each prediction head, a fully connected layer with the sigmoid-activation function was used. 

We pre-trained the parts of the multimodal transformer architecture. The audio encoder was pre-trained with masked prediction of hidden units \cite{hsu_taslp2021} using over 20K hrs of Japanese speech. The text encoder was pre-trained with a masked language-modeling task \cite{devlin_naacl2019} using over 100G tokens of text. The visual encoder was pre-trained with a still-image-based facial-expression-recognition task using RAF-DB \cite{p3} and AffectNet \cite{p4} datasets. Note that these pre-trained parameters were not frozen in the following main training. After the pre-training, all parameters in each model were trained. The mini-batch size was set to 8, and the dropout rate in the transformer blocks was set to 0.1. We used RAdam \cite{liu_iclr2020} for optimization. The training steps were stopped on the basis of early stopping using the validation dataset. We trained all models with one NVIDIA A6000 GPU.

\subsection{Evaluation metrics}  We evaluated task-specific models and a joint model in terms of Pearson's correlation coefficient and accuracy. The accuracy was computed in the same manner as with ChaLearn first impression \cite{lopez_eccv2016,escalante_ijcnn2017}. The accuracy for the $k$-th personality trait against the $D$ test samples is defined as 
\begin{equation}
    {\rm Accuracy}_k = 1 -  \frac{1}{D} \sum_{d = 1}^{D} |\hat{y}_k^d - y_k^d| ,
\end{equation}
where $\hat{y}_k^d$ and $y_k^d$ are the ground-truth and predicted scores of the $k$-th personality trait for the $d$-th test sample. Note that the scores were normalized in the range of [0, 1].

\subsection{Results}
 Tables 1 and 2 show the multimodal apparent-personality-trait recognition performance for the Big Five and HEXACO, respectively. The experimental results show that audio features were more effective than visual features, and the visual features are comparatively effective in recognizing {\it Agreeableness} in the Big Five, {\it Emotionality} and {\it Agreeableness} in the HEXACO. In addition, combining audio, text, and visual inputs was effective for both the Big Five and HEXACO. This indicates that a multimodal transformer architecture with pre-trained encoders was effective in integrating multimodal information. The experimental results also show that the joint model outperformed the task-specific models for Big Five and HEXACO in most cases. This suggests that we can promote robustness to being aware of various multimodal human behaviors by explicitly taking into account the relationships between the Big Five and HEXACO.
The highest performance was achieved by the joint model with audio, visual, and text inputs for the Big Five and HEXACO evaluation. The automatic recognition performance competed with human evaluation performance.

Table 3 shows the correlations between the Big Five and HEXACO for the human and automatic evaluation using the joint model with audio, visual, and text inputs. The results of the human evaluation show that the correlations between the Big Five and HEXACO were as expected for the characteristics considered highly correlated between the Big Five and HEXACO. {\it Honesty-Humanity} in the HEXACO did not correlate highly with any of the traits in the Big Five. These results indicate that our experimental setups using our newly annotated dataset were convincing for evaluating apparent Big Five and HEXACO traits. Next, the results of the automatic evaluation show that the correlations were higher not only for the characteristics considered highly correlated between the Big Five and HEXACO but also for other traits. This is because correlations between the Big Five and HEXACO were excessively captured during the model training. While we could achieve recognition performance equivalent to human performance, we could not reproduce the way humans perceive impressions. Bringing the correlations between the Big Five and HEXACO closer to human evaluation will be a future challenge.

\begin{table}[t!]
  \centering
   \caption{Correlation matrix between Big Five and HEXACO}
  \begin{tabular}{l|rrrrrr} \toprule
      \multicolumn{7}{c}{Human evaluation} \\
      & {\it H} & {\it E} & {\it X} & {\it A} & {\it C} & {\it O} \\ \midrule
      {\it O} &  0.134 & -0.155  & 0.479  & 0.378 & 0.539  & {\bf 0.797}  \\
      {\it C} & 0.432 & 0.066  & 0.170  & 0.464 & {\bf 0.837}  & 0.518 \\
      {\it E} & -0.363 & -0.301  & {\bf 0.937}  & 0.114 & -0.05  & 0.355  \\
      {\it A} & 0.362 & 0.179  & 0.462  & {\bf 0.7621} & 0.430  & 0.558  \\ 
      {\it N} & 0.078 & -0.517  & 0.643  & 0.424 &0.266  & 0.438 \\ \midrule
      \multicolumn{7}{c}{Automatic evaluation with the proposed joint model} \\
      & {\it H} & {\it E} & {\it X} & {\it A} & {\it C} & {\it O} \\ \midrule
      {\it O} &  0.299 & -0.002  & 0.564  & 0.515 & 0.805  & {\bf 0.947}  \\
      {\it C} & 0.652 & 0.333  &0.297  & {\bf 0.727} & {\bf 0.924}  & {\bf 0.850} \\
      {\it E} & -0.553 & -0.247  & {\bf 0.984} & 0.105 & -0.07  & 0.363 \\
      {\it A} & 0.500 & 0.554  & 0.529  & {\bf 0.921} & 0.567  & {\bf 0.7183} \\
      {\it N} & -0.302 & -0.440  & {\bf 0.833}  &0.224 & 0.189  & 0.524 \\ \bottomrule
    \end{tabular}
\end{table}

\section{Conclusion}
We demonstrated the first investigation of automatically recognizing observer-perceiving HEXACO traits from multimodal human behavior. We also introduced a novel joint modeling method of Big Five and HEXACO to consider the relationships between them. The experimental results demonstrated the effectiveness of the proposed joint modeling approach, showing improved recognition performance for both the Big Five and HEXACO traits.  Future work includes bringing the correlations between the Big Five and HEXACO closer to human evaluation.

\bibliographystyle{IEEEtran}
\bibliography{mybib}

\end{document}